# Generative KI für TA

*Wolfgang Eppler, Reinhard Heil, ITAS, KIT*

**Abstract**[1]

Viele Wissenschaftler*innen nutzen generative KI in ihrer wissenschaftlichen Arbeit. In der Technikfolgenabschätzung (TA) arbeitende Menschen sind hierbei nicht ausgenommen. Der Umgang der TA mit generativer KI ist ein doppelter: Einerseits wird generative KI für die TA-Arbeit genutzt, andererseits ist generative KI Gegenstand der TA-Forschung. Der folgende Beitrag geht, nachdem er kurz das Phänomen generativer KI umrissen und Anforderungen für deren Einsatz in der TA formuliert hat, ausführlich auf die strukturellen Ursachen der mit ihr verbundenen Probleme ein. Generative KI wird zwar ständig weiterentwickelt, die strukturell bedingten Risiken bleiben jedoch bestehen. Der Artikel schließt mit Lösungsvorschlägen und kurzen Hinweisen zu deren Umsetzbarkeit sowie einigen Beispielen für den Einsatz generativer KI in der TA-Arbeit ab.

## 1. Phänomene der generativen KI

Ein herausragendes Merkmal generativer KI (im Folgenden synonym verwendet mit Chatbot und multimodalem Sprachmodell LLM) ist, dass sie es ermöglicht, mit Maschinen schriftlich oder mündlich in natürlicher Sprache zu kommunizieren. Chatbots, wie ChatGPT, Gemini, LLama oder Claude, die auf sogenannten großen Sprachmodellen basieren, lassen sich sehr einfach, ohne Einarbeitung nutzen. Man kann mit ihnen plaudern (Tuschling et al. 2023), ihnen Fragen stellen, sie können Texte zusammenfassen und aus Eingaben, sogenannten Prompts, Texte und Bilder in beliebigen Formen und Stilen erzeugen (Yin et al. 2024). Die Anwendungen sind aber nicht auf natürliche Sprache beschränkt, sie können u.a. auch Programmcode erzeugen. Mittlerweile führen Chatbots (als Agenten, beispielsweise durch Large Action Models (Hille 2025) Aktionen aus, sie können bspw. im Internet suchen oder Unternehmensdaten verarbeiten (Gao et al. 2024).

Die LLM sind jedoch an erster Stelle Sprachmodelle und keine Wissensmodelle. Sie geben nicht ein einmal gelerntes Wissen wieder, sondern generieren, anhand der während des Trainings erkannten statistischen Zusammenhänge, neue Texte, die faktisch richtig sein können, aber nicht müssen. Sie können Fehlinformationen (*misinformations, desinformations, hallucinations* (Deng et al. 2024)) erzeugen, was ihre Vertrauenswürdigkeit untergräbt. Da die

---

[1] 

Texte, mit denen Sprachmodelle trainiert werden, alle Arten von destruktiven, hasserfüllten und toxischen Aussagen enthalten, finden sich Vorurteile, Voreingenommenheit, verzerrte Ansichten (*bias*) auch in der Ausgabe generativer KI wieder, oft in verstärktem Maße (Parrish et al. 2022). Um das Erzeugen solcher unerwünschten, teils von Nutzenden gezielt angeregten, Ausgaben (Fehlausrichtung (*misalignment*)) zu verhindern, werden Korrekturmaßnahmen durchgeführt. Dieses sogenannte *alignment* bleibt jedoch nicht ohne Folgen, es kann selbst zu Verzerrungen führen (Askell et al. 2021). Generative KI ist intransparent, es ist zumeist nicht möglich, genau nachzuvollziehen, wie eine Ausgabe zustande kommt. Es gibt zwar eine große Zahl von Forschungsansätzen, die versuchen, KI erklärbar zu machen, diese können aber bisher die in sie gesetzten Erwartungen nicht erfüllen (e.g. Wu et al. 2024, Renftle et al. 2022).

## 2. Strukturelle Ursachen von Problemen derzeitiger KI

Eine paradigmatische Aufgabe der TA ist die Politikberatung (Grunwald 2023). Sie erfordert verlässliche, kluge, situationsgerechte Lösungen. Um dies zu gewährleisten, müssen die, von generativer KI erzeugten, Ausgaben verständlich, nachvollziehbar, kontrollierbar, kohärent, diskriminierungsfrei und erklärbar sein sowie Quellen und verwendete Daten offengelegt werden.

Im Folgenden werden die strukturellen Ursachen von Problemen derzeitiger KI untersucht, um besser einschätzen zu können, wo generative KI in der TA eingesetzt werden kann und wo nicht. Es lassen sich zumindest acht strukturelle Ursachen derzeitiger KI-Probleme aufweisen.

*(1) Datenqualität*

Der größte Teil der für das Training der KI verwendeten Daten kommt aus dem Internet (siehe (Liu et al. 2024b)). Die Qualität dieser Daten ist oft schlecht, da u.a. viele synthetische Daten (von Maschinen erzeugt, Systemprotokolle, …) enthalten sind (Shumailov et al. 2024), Daten mehrfach vorkommen (Carlini et al. 2023, Deng et al. 2024) oder, gerade bei neuen gesellschaftlichen Entwicklungen, widersprüchlich sind (Augenstein et al. 2024). Zudem ist es praktisch unmöglich Datensätze dieser Größenordnung auf sachliche Richtigkeit zu prüfen. Neben den Internetdaten werden zum Training auch digitalisierte Bücher und Zeitschriften genutzt. Die Daten sind auf keinen Fall gleichverteilt, sie enthalten Lücken, die dem Anwender verborgen bleiben. Verteilungsverschiebungen[2] führen mit dazu, dass Modelle in neuen

---

[2] Die Menge der Trainingsdaten ist nicht gleichverteilt. Sie weist an verschiedenen Stellen des Datenraums Häufungen bzw. Löcher auf. Wird ein Modell häufig in einem Bereich angewandt, der von den Daten schlecht abgedeckt ist (Datenloch), weisen die Anwendungsdaten gegenüber den Trainingsdaten eine verschobene Verteilung auf.

Situationen falsch generalisieren (Liu et al. 2024a, Vafa et al. 2024). Die Anbieter generativer Modelle bereinigen fast immer die Originaldaten und setzen Clickworker zum Annotieren ein. Wie dies genau geschieht und welche Anpassungen vorgenommen werden, bleibt meist im Dunkeln (Rohde 2024, Albalak et al. 2024).

*(2) Fehlausrichtung*

Zwar verfügen auf generative KI basierende Chatbots über beeindruckende sprachliche Fähigkeiten, sie geben aber immer wieder ethisch zweifelhafte Antworten. Um dem zu begegnen, wurden verschiedene, aufeinander aufbauende Fehlausrichtungskorrekturen (*alignment*) entwickelt. Als erstes ist das Fine-Tuning zu nennen. Fine-Tuning bedeutet, dass das Training anhand großer Datenmengen (Pre-Training) ergänzt wird, indem das Sprachmodell mit annotierten Anfrage-Antwortpaaren trainiert wird (Ouyang et al. 2022). Diese Anfrage-Antwortpaare sind im Vergleich zu den Trainingstexten des Pre-Trainings anwendungsbezogener, von Menschen explizit bewertet (*supervised*) und von deutlich geringerer Anzahl als die Texte des Pre-Trainings. Ziel ist es, toxische Inhalte, also Diskriminierungen, Hassreden, destruktive Handlungen, usw. zu erkennen, abzumildern oder zu korrigieren. Da die zusätzlichen Trainingsdaten stärker auf die alltägliche Nutzung der ChatBots zugeschnitten sind und für entsprechende Anwendungen die Datenlage verbessern, sollen sie zugleich die Anzahl der Fehlinformationen reduzieren. Ähnlich wie das Fine-Tuning dient das *Instruction Tuning* dazu, die Modell-Antworten auf Alltagsanfragen zu verbessern. Eine ganz andere Art von Ausrichtung ist durch die Verwendung von Prompt-Erweiterungen möglich. Der System Prompt teilt dem generativen Modell vor jeder Benutzeranfrage bereits einige Informationen mit, beispielsweise, wie das Modell heißt, aber auch, wie das Modell in bestimmten Situationen reagieren soll. Beispielsweise wird dem System so mitgeteilt, dass es die Beantwortung einer Aufgabe Schritt für Schritt angehen soll (*Chain of Thought (CoT) Prompt* (Wei et al. 2023)), um so seine eigenen Ausgaben in den Lösungsweg mit einzubeziehen. Fine-Tuning und Prompts können von den Nutzenden nochmals zusätzlich gestaltet werden, um das generative Modell noch stärker auf ihre Anforderungen zuzuschneiden.

Diese Alignment-Methoden haben sehr zum Erfolg der generativen KI beigetragen. Sie haben aber die Zuverlässigkeit der Modelle nicht genügend erhöht, sie können sie teilweise sogar verringern. So können durch Fine-Tuning Inhalte, die durch Pre-Training gelernt wurden, überschrieben und vergessen werden (*catastrophic forgetting* (Ven et al. 2024)), sich toxische Inhalte einschleichen oder eine ausgewogene Datenverteilung aus dem Gleichgewicht gebracht werden (Qi et al. 2023). Woran liegt diese Resistenz gegenüber allen Korrekturmaßnahmen? Ein Grund ist, dass Menschen und ML-Modelle oft nicht die gleichen Merkmale nutzen, um Kategorien zu bilden (*Proxy-Effect* (John et al. 2024)). So kommt es vor, dass eine KI generalisierende Merkmale herausbildet, die nur aufgrund von Randeffekten, wie z.B. ein Wasserzeichen, eine Unschärfe im Bild oder ein belanglos eingeflochtener Nebensatz, die für einen Menschen unerheblich sind, zum gewünschten Ergebnis führen. Dies führt auch dazu, dass KI unerwartete oder andere Fehler macht als Menschen.

*(3) Kontext-Inhalt-Herausforderung*

Es gibt ein Spannungsverhältnis zwischen der Benutzereingabe (Kontext) und dem, was ein Sprachmodell während des Trainings gelernt hat, da die Benutzereingabe Aussagen enthalten kann, die dem gelernten Inhalt widersprechen. Die spannende Frage ist dann, welcher Anteil sich durchsetzt (Min et al. 2022, Dai et al. 2023), Kossen et al. 2024). Bekannt ist, dass das Anführen von Beispielen im Eingabe-Kontext, sogenanntes *In-Context Learning* oder *Few Shot Prompting*, die Ausgabe-Qualität eines Sprachmodells stärker verbessern kann als das Fine-Tuning (Brown et al. 2020). Das ist besonders bei Aufgaben einsichtig, die eine Ausgabe in Tabellenform erlauben und als annotierte Eingabe die drei oder vier ersten Zeilen der Tabelle explizit mitgeteilt bekommen. Für den Rest der Tabelle wird nur noch die Eingabe vorgegeben und die Ausgabe vom Sprachmodell ergänzt. Diese Vorgehensweise suggeriert, dass von außen die Form vorgegeben wird, während der Inhalt durch Informationen des Modells bestimmt wird. Es zeigt sich jedoch auch, besonders bei umfangreicherem Kontext, dass auch Inhalte des Kontextes in die Antwort übernommen werden – unabhängig davon, ob sie richtig oder falsch sind (*sycophancy* (dt. Kriecherei)). Die Intransparenz der Daten wie auch die Komplexität der Modellarchitektur sorgen dafür, dass nicht vorhergesagt werden kann, in welchen Fällen sich der Kontext und wann sich der Inhalt durchsetzt. Das Ergebnis ist unbestimmt.

*(4) Reproduktion*

Der Trainingsprozess geschieht nicht fortlaufend, nicht während der Dialoge mit den Chatbots, sondern wird einmalig mit riesigen Datensätzen durchgeführt und kann Wochen dauern. Deshalb sind große Sprachmodelle außerstande, eigene Erfahrungen zu machen. Sie können nicht kontinuierlich lernen. An Trainingsalgorithmen, die dies ermöglichen, wird intensiv geforscht (Shi et al. 2024))

Aktuelle gesellschaftliche Veränderungen und menschliche Erfahrungen sickern so erst allmählich in Texte ein, mit denen dann eine KI trainiert wird. Die KI hinkt also der gesellschaftlich-technischen Entwicklung im Vergleich zum Menschen hinterher. So ergibt sich eine zeitliche Kluft zwischen einer aktuellen Benutzer-Anfrage und dem Aufkommen einer Veränderung mit den nacheinander folgenden Schritten:

- gesellschaftliche Veränderungen, die noch nicht in Sprache gefasst wurden.
- textliche Verarbeitung eines Themas in der Gesellschaft (Informationsquelle)
- Festlegung des Korpus mit Trainingsdaten
- Release-Zeitpunkt des Sprachmodells
- Benutzeranfrage an eine KI

Die KI braucht Texte schreibende Menschen für ihre Reproduktion. Synthetische Texte können diese Grundlage nicht liefern. Das von (Harnad 1990) zuerst beschriebene Grounding-Problem für symbolische KI ist von Bender und Koller (2020) für die generative KI bestätigt worden. U.a. die Arbeiten der Philosophen Habermas (1981) und Brandom (1994) zeigen, dass für einen Realitätsbezug kontinuierliches Lernen und verantwortungsbewusste Setzungen unabdingbar sind. Dies können Sprachmodelle bisher nicht leisten.

*(5) Fehlen sozialer Perspektiven*

Von einer KI erwarten wir, nachdem wir eine falsche Antwort korrigiert haben, dass das Modell die Korrektur annimmt und das nächste Mal berücksichtigt. Unter *(4) Reproduktion* konnten wir sehen, dass dazu kontinuierliches Lernen notwendig wäre, wozu heutige Modelle nicht in der Lage sind. Eine andere Situation stellt sich ein, wenn bei der Bewertung einer Handlung die Meinungen auseinander gehen. Die Klärung erfolgt dann nicht mehr direktiv wie in einem asymmetrischen Eltern-Kind- oder Lehrer-Schüler-Verhältnis, sondern symmetrisch durch einen Diskurs, in dem über den Austausch von Argumenten geklärt wird, wer Recht hat. Um einen solchen Diskurs über Wahrheit zu führen, muss man in der Lage sein, verschiedene soziale Perspektiven einnehmen zu können (Habermas 1981, (Brandom 1994). Die Perspektive generativer KI ist funktional. Ihr fehlt die normative Komponente, die einen menschlichen Gesprächspartner auszeichnet. Warum setzt ein Wahrheitsdiskurs solche sozialen Perspektiven voraus? Eine zweite Perspektive generalisiert die Meinung einer einzigen Perspektive, die dadurch intersubjektiviert wird. Fehlt die zweite Perspektive, fehlt ein normatives Korrelat, das die richtige Verwendung eines Begriffs bestätigt. Die soziale Perspektive ermöglicht es, nicht nur zwischen bloßen Behauptungen und begründbaren

Aussagen zu unterscheiden, sondern die Sprecher übernehmen zudem Verantwortung für das von ihnen Gesagte, sie gehen Verpflichtungen ein.

*(6) Weltmodell*

Wenn im Rahmen der generativen KI von Weltmodell gesprochen wird, ist damit zumeist ein physikalisches Weltmodell mit körperlichen Fähigkeiten in einer raumzeitlichen Welt mit physikalischen Gesetzmäßigkeiten gemeint (Xu et al. 2024). Die körperlichen Fähigkeiten sind im Gegensatz zu den in den Sprachmodellen zweifellos vorhandenen sprachlichen Fähigkeiten zu sehen. Demgegenüber gibt es auch phänomenologische Weltmodelle, welche eine Welt mit Erfahrungen und Wahrnehmungen abbilden. Dieser Aspekt wurde unter *(4) Reproduktion* behandelt. Psychologische Weltmodelle umfassen die mentale Welt mit unserem Alltagsverstand und Schlussfolgern (s.u. *(7) Reasoning*). Bei allen drei verschiedenen Arten von Weltmodellen spielen Veränderungen eine Rolle, in Form von Bewegungen (physikalisch), Erfahrungen (phänomenologisch) oder Lernen (psychologisch).

Die zurzeit diskutierten Weltmodelle für die KI lassen sich in drei aufeinanderfolgenden Stufen einteilen: Videogeneratoren, Simulationsmodelle und reale Weltmodelle. Ein vielbeachteter Videogenerator ist *SORA* von OpenAI (Brooks et al. 2024). Durch Auswertung vieler Videos gelingt es, die in Videos beobachteten Bewegungen zu generalisieren – allerdings ohne die dahinterliegenden physikalischen Gesetze erfahren zu haben. In vielen mit *SORA* erzeugten Videos können Unstimmigkeiten beobachtet werden wie eine fehlende Objektpersistenz und physikalische Inkohärenzen über längere Zeit. Simulationsmodelle kommen in Computerspielen vor, in denen Agenten in einer künstlichen Welt erfundenen Gesetzen unterworfen sind. Damit wirkt die künstliche Welt auf die Agenten zurück, was bei Videogeneratoren fehlt. Allerdings fehlen der künstlichen Welt die Unvorhersehbarkeiten einer realen Welt. Dies erfahren Roboter, die vorher in Simulationen trainiert wurden und anschließend der realen Welt ausgesetzt werden. Es gibt zurzeit aber keine Trainingsalgorithmen aus der generativen KI, die das leisten (siehe *(4) Reproduktion* mit fehlendem kontinuierlichem Lernen).

Fehlt ein realistisches Weltmodell, wird eine generative KI immer wieder inkohärente Ausgaben erzeugen und mit raumzeitlichen Zuordnungen Schwierigkeiten haben.

*(7) Reasoning*

*Reasoning* (Schlussfolgern mit gesundem Menschenverstand) besitzt ein weites Bedeutungsspektrum. Sun et al. (2023) (siehe Bild 1) zeigen, wie vielfältig dieser Begriff gebraucht wird. Generative KI schneidet, was verlässliches Schlussfolgern angeht, noch ziemlich schlecht ab. Als Gründe für den fehlenden Alltagsverstand können u.a. ein fehlendes physikalisches Weltmodell (siehe *(6) Weltmodell*) und fehlende körperliche und kontinuierliche Erfahrungen genannt werden. Die Reflexion generativer KI mithilfe von Zwischenschritten (Chain of Thought mit "*hidden CoT prompt*", siehe *(2) Fehlausrichtung*) soll eine weitere *soziale Perspektive (5)* implementieren, um so logisches Schließen verlässlicher zu machen. Angeblich sind die großen Sprachmodelle damit auf einem guten Weg, besser zu werden. Ein Durchbruch wurde angeblich mit ChatGPT-4o1 geschafft. Kurz nach Einführung des neuen Systems wurde jedoch bereits aufgezeigt, welche Lücken sich beim zielgerichteten Schlussfolgern zeigen. Das komplizierte und falsche Schließen (z.B. bei einem abgewandelten Flussüberquerungsrätsel) ist erklärbar, wenn man davon ausgeht, dass vergleichbare Aufgaben bereits im Trainingsdatenset vorhanden waren und deren Struktur fälschlich auf neue Aufgaben übertragen wurden (Mirzadeh et al. 2024). Wird die Aufgabe in einem wesentlichen Detail abgewandelt, geht das Schließen fehl.

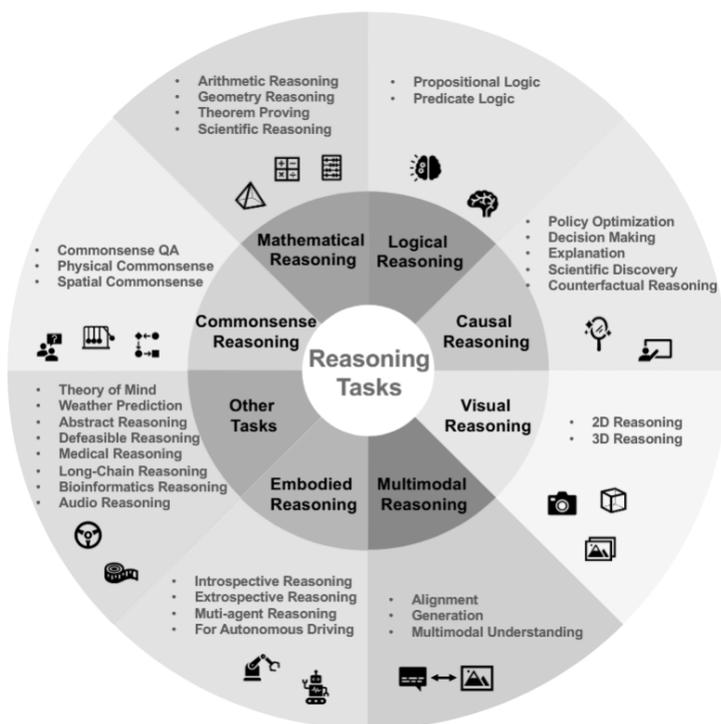

*Bild 1: Reasoning: Verschiedene Arten von Schlussfolgern. Quelle: Sun et al. 2023*

### 3. Einsatzmöglichkeiten generativer KI für die TA

Aufgrund der angeführten Schwächen ist es ratsam, generative KI sehr bedacht und nur dann einzusetzen, wenn man in der Lage ist, die Ergebnisse zu überprüfen. Für die Technikfolgenabschätzung ist die KI sowohl Untersuchungsgegenstand als auch mögliches

Werkzeug. Im Folgenden werden, ohne Anspruch auf Vollständigkeit, in der TA anfallende Aufgaben daraufhin überprüft, inwiefern sie angesichts der zuvor herausgearbeiteten Herausforderungen mit Hilfe generativer KI bearbeitet werden können.

Das sogenannte Horizon-Scanning (Hungerland 2025), wie es unter anderem am Büro für Technikfolgenabschätzung beim Deutschen Bundestag (TAB) durchgeführt wird, eignet sich gut um die Möglichkeit und Grenze des Einsatzes generativer KI in der TA zu verdeutlichen. Am Anfang des Horizon-Scannings, wie jedes TA-Projektes, steht die Informationssammlung mittels einer strukturierten Suche nach und Auswertung von Quellen. Ausgewertet werden Printmedien, Onlinequellen (Webseiten, Blog, Datenbanken etc.), wissenschaftliche Publikationen, Reports/Berichte, es werden aber auch Expert/innen und Laien befragt. Es gibt mittlerweile eine Unzahl an auf generativer KI basierender KI-Tools, die die Auswertung solcher Quellen erleichtert. Alle diese Tools sind jedoch unzuverlässig. Ob beispielsweise eine automatisch erstellte Literaturübersicht alle relevanten Quellen enthält, sachlich richtig ist, keine Übertreibungen, erfundenen Quellen und anderer sogenannte Halluzinationen enthält muss geprüft werden, ebenso automatisiert erstellte Interviewtranskripte. Generative KI kann zwar das Schreiben von Überblicken erleichtern, aber auf keinen Fall vollständig übernehmen. Zur Informationssammlung gehört auch die Integration und Anreicherung der gefundenen Quelle, qualitative Inhaltsanalysen, sowie Interpretation, Diskussion und Validierung der gefundenen Informationen. Generative KI kann hier insbesondere dazu genutzt werden, um bei der Clusterung von Informationen zu helfen und Anregungen zu geben. Solange entsprechenden Vorschlägen nicht blind vertraut wird, kann dies die Informationsauswertung und -aufbereitung erleichtern. Gleiches gilt für den auf die Informationssammlung folgenden Schritt des Horizon-Scanning, die Ideenaufbereitung. Ihre eigentlichen Vorteile kann die generative KI aber insbesondere bei der sprachlichen und graphischen Aufbereitung der Ergebnisse ausspielen: Erzeugung von Fließtexten aus Aufzählungen und umgekehrt, Erstellung von Folien und graphischen Darstellungen, Textkorrektur und -erzeugung, Übersetzungsunterstützung etc. Selbstverständlich müssen auch bei diesem Schritt alle automatisch generierten Ergebnisse auf Korrektheit geprüft werden.

## 4. Konklusion

Grob zusammengefasst sollte derzeit die Anwendung generativer KI in der TA darauf beschränkt werden, sie als Ideengeber und als Unterstützung zu nutzen. Ungeprüft dürfen die

Ergebnisse nicht verwendet werden, man darf sich nicht auf sie verlassen. Damit unterscheidet sich der Einsatz generativer KI in der TA kaum von ihrem Einsatz in anderen wissenschaftlichen Bereichen.

Wir sollten aber nicht vergessen, dass es in der TA es um mehr geht als möglichst schnell möglichst viele Informationen zu sammeln und zusammenzufassen. Wir leiden unter akademischem FOMO (fear of missing out), wir haben ständig Angst, etwas Wichtiges zu übersehen. Generative KI wird oft als Retter in der Not dargestellt, als einzige Möglichkeit, der Informationsflut Herr zu werden. Das mag zwar stimmen, aber die Idee, dass über mehr Informationen zu verfügen immer besser ist, ist von vorneherein falsch. Mehr Informationen tragen oft gar nicht nennenswert zu einem besseren Verstehen bei. Intensives Lesen ist ein wichtiger Teil des Verstehens, ebenso wie das Zusammenfassen und Schreiben. Was uns Maschinen tatsächlich niemals werden abnehmen können, ist zu verstehen und wir müssen darauf achten, dass wir die zum Verstehen notwendigen Fähigkeiten nicht aus Zeitdruck und Bequemlichkeit verlernen.